\documentclass[conference]{IEEEtran}
\IEEEoverridecommandlockouts
\usepackage{cite}
\usepackage{caption}
\usepackage{amsmath,amssymb,amsfonts,amsthm}
\usepackage{algorithmic}
\usepackage[ruled,vlined]{algorithm2e}
\usepackage{booktabs}
\usepackage{graphicx}
\usepackage{float}
\usepackage{graphicx}
\usepackage{textcomp}
\usepackage{color}
\usepackage{threeparttable}
\usepackage{makecell}
\usepackage{multirow}
\usepackage{xcolor}
\usepackage{tikz}
\usepackage{subcaption}
\usepackage{adjustbox}
\usetikzlibrary{positioning, shapes.geometric}
\def\BibTeX{{\rm B\kern-.05em{\sc i\kern-.025em b}\kern-.08em
    T\kern-.1667em\lower.7ex\hbox{E}\kern-.125emX}}

\begin{document}

\title{EDCSSM: Edge Detection with Convolutional State Space Model\\

}
\author{\IEEEauthorblockN{Qinghui Hong, Haoyou Jiang, Pingdan Xiao, Sichun Du, Tao Li$^*$}

{\footnotesize \textsuperscript{}}
\thanks{

This paper was supported in part by National Natural Science Foundation of China under Grant 62234008, 62371186; in part by Huxiang young talentsunder Grant 2023RC3103; in part by the Natural Science Foundation ofHunan Province under Grant 2023JJ30168, 2022JJ30160, 2021JJ40111; inpart by the National Key R\&D Program of China Grant 2022YFB3903800).

}
}

\maketitle

\begin{abstract}
Edge detection in images is the foundation of many complex tasks in computer graphics. Due to the feature loss caused by multi-layer convolution and pooling architectures, learning-based edge detection models often produce thick edges and struggle to detect the edges of small objects in images. Inspired by state space models, this paper presents an edge detection algorithm which effectively addresses the aforementioned issues. The presented algorithm obtains state space variables of the image from dual-input channels with minimal down-sampling processes and utilizes these state variables for real-time learning and memorization of image text. Additionally, to achieve precise edges while filtering out false edges, a post-processing algorithm called wind erosion has been designed to handle the binary edge map. To further enhance the processing speed of the algorithm, we have designed parallel computing circuits for the most computationally intensive parts of presented algorithm, significantly improving computational speed and efficiency. Experimental results demonstrate that the proposed algorithm achieves precise thin edge localization and exhibits noise suppression capabilities across various types of images. With the parallel computing circuits, the algorithm to achieve processing speeds exceeds 30 FPS on 5K images.
\end{abstract}


\section{Introduction}
Edge detection is one of the most fundamental tasks in the field of computer graphics and is a crucial step for advanced tasks such as object detection, image segmentation, and 3D reconstruction\cite{ref1}–\hspace{-0.005em}\cite{ref5}. However, achieving precise and comprehensive edge detection is challenging. This difficulty arises because edges in images can degrade due to factors such as lighting conditions, image noise, and scene complexity. 

To overcome these challenges, traditional methods have employed various handcrafted features and thresholding techniques to enhance the algorithm's edge perception capabilities \cite{ref6}\cite{ref7}. However, these methods exhibit poor noise suppression and suffer from performance degradation in precise edge localization when dealing with more challenging and complex images. To further improve performance, many advanced functional units have been developed or borrowed from other domains for edge detection \cite{ref8}–\hspace{-0.005em}\cite{ref11}. While these methods achieved better performance, they fail to effectively utilize the contextual information of the image, thereby struggling to filter out false edges and exhibiting limited noise suppression capabilities.

Utilizing the powerful generalization capabilities of machine learning models for image edge detection can effectively leverage the contextual information of images to achieve comprehensive edge detection and noise suppression. For instance, built on top of the ideas of fully convolutional neural networks and deeply supervised nets, HED detects complete edges of objects from complex images \cite{ref12}. CEDN achieves remarkable detection performance through a fully convolutional encoder-decoder network \cite{ref13}. Recently, other machine learning methods have also been employed to improve edge detection performance\cite{ref14}–\hspace{-0.005em}\cite{ref17}. However, the inherent multi-layer down-sampling structures of deep learning methods lead to feature loss, causing the detected edges to be thick and the omission of edge features of small-scale objects, which lead to these methods unsuitable for high-precision, full-size edge detection tasks. Additionally, the high pre-training cost is also a hindrance to miniaturization.

In summary, we believe the community still needs an edge detector that can accurately detect edges at all scales, efficiently utilize image contextual information to suppress noise edge and is optimized for resource-constrained devices.

\begin{figure}[htbp]
    \centering
    \begin{subfigure}[b]{0.46\textwidth}
        \centering
        \includegraphics[width=\textwidth]{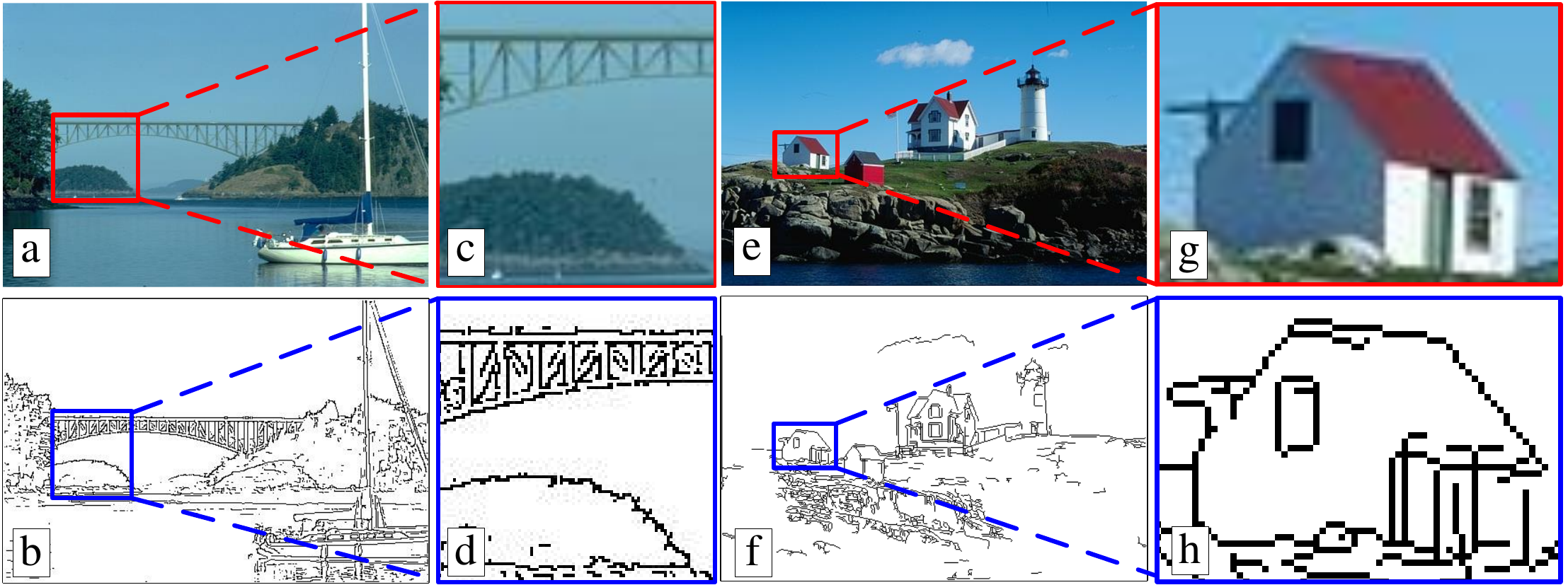}
    \end{subfigure}
  \caption{
\textbf{Examples of edge detection}. 
Our method, EDCSSM, learns and extracts precise edges at all scales in images through state space variables. (a, e): Input images from BSDS500 \cite{ref_dataset_BSDS500}. (b, f): Detected edges by EDCSSM. (c, d, g, h): Zoomed-in patches.}
\label{fig1} 
\end{figure}

In this work, we attempt to meet this need by applying the Mamba architecture for edge detection. The Mamba architecture \cite{ref18}–\hspace{-0.005em}\cite{ref24}, a recent success in the field of deep learning, possesses ultra-long contextual memory and minimal down-sampling stages, making it highly promising for achieving the desired outcomes. However, several challenges need to be addressed. Firstly, edge detection is typically based on convolution operations, whereas the Mamba architecture uses the state space model based on matrix operations. To the best of our knowledge, there are unavailable state space models based on convolution operations. Therefore, designing a convolution-based state space model suitable for edge detection is crucial. Secondly, as shown in Fig. \ref{fig1} (e), filtering out unwanted edges from complex images (such as the exposed soil under the lighthouse and the rubble on the shore in the lower right corner) and accurately extracting the edges of small objects (such as the houses on the mountain, the lighthouse, and the flag) is challenging.
 Therefore, it is necessary to design appropriate convolution operators that can extract the desired state variables. Lastly, to enhance the algorithm's performance, new post-processing strategies, in addition to conventional post-processing procedures, need to be designed to meet the specific post-processing requirements of the algorithm. 

\begin{figure*}[htbp]
  \centering
  \includegraphics[width=\textwidth]{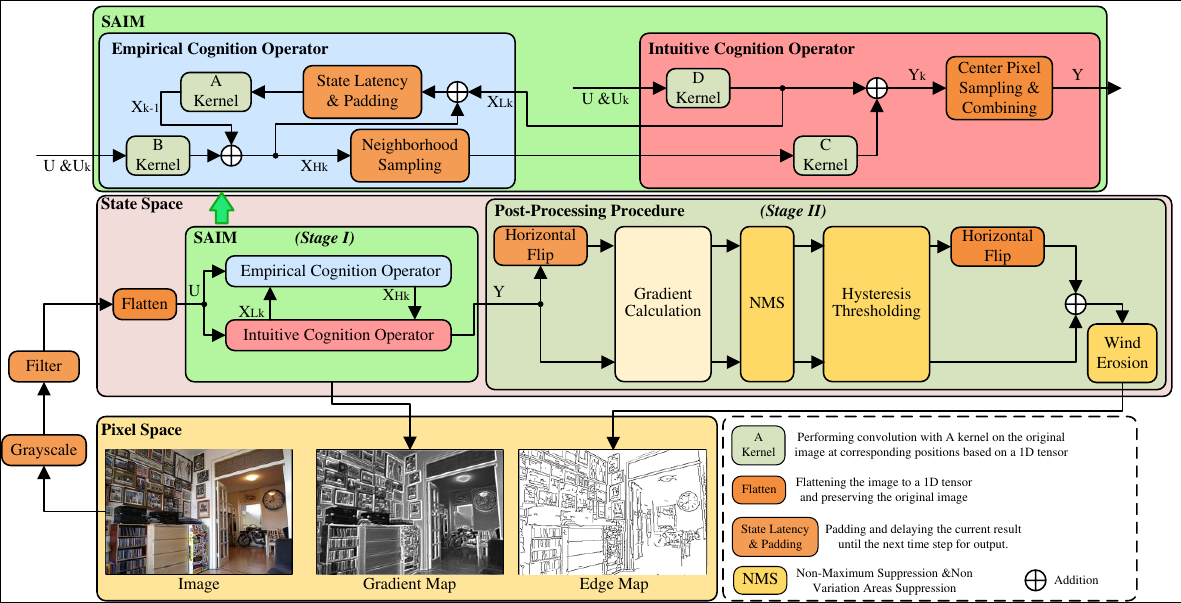}
  \caption{The framework of the proposed EDCSSM.}
  \label{fig2}
\end{figure*}

To address the above issues, we modify the original state space equations and design a comprehensive framework (Fig. \ref{fig2}) named Edge Detection with Convolutional State Space Model for  (EDCSSM), aiming at perceiving edges at all scales and filtering out false edges. In $stage$ $I$ of Fig. \ref{fig2}, the image is split into tensors and sequentially input into the State Space Model (SAIM), which continuously learns edge information from previous text and feeds the results back to the current input. In $stage$ $II$, we first process the gradient maps generated by SAIM through conventional post-processing steps. The resulting edge maps are then passed through a dedicated post-processor called ``Wind Erosion," which includes the following eight steps: 1) find boundary, 2) process long edges, 3) split edges, 4) clear edges, 5) restore junctions, 6) restore protected edges and 7) restore boundary. Through these efforts, EDCSSM can accurately capture edges at all scales and effectively filter out false edges (Fig. \ref{fig1}). Finally, to improve the processing speed and efficiency of EDCSSM, we design specialized parallel computing circuits to handle the most computationally intensive parts. Our contributions include:
\begin{itemize}
\item Proposing a novel state space model framework for edge detection (EDCSSM) that effectively captures edges at all scales and filters out false edges.
\item Developing a special post-processing procedure named ``Wind Erosion", consisting of seven specific steps to refine and clean the detected edges.
\item Implementing specialized parallel analog-computation circuits to enhance the processing speed and efficiency of the proposed method.
\end{itemize}

\section{Related Work}
\subsection{Edge Detection}
Edge detection aims to precisely delineate boundaries
and visually salient edges from various image. Due to the high diversity of its content, many conventional algorithms have been developed mainly based on assumptions and hand-crafted features\cite{ref6}\cite{ref7}, which are unsuitable to more challenging practical applications. In recent years, new methods have achieved notably superior performance over traditional counterparts by integrating techniques from multiple domains. These methods can be categorized into two types: advanced feature detection techniques and machine learning-based detection methods. The advanced feature detection techniques improve edge perception capabilities by introducing image feature analysis techniques from other fields. For example, Xu et al. combined the edge detection algorithm with the mathematical morphology nonlinear filtering to suppress noise and enhance edges\cite{ref25}. Shekar et al. applied Taylor’s Expansion Theory to suppress image details and enhance object contours\cite{ref26}. MSCNOGP utilized multi-scale closest neighbor operator with grid partition technique to improve detection accuracy and noise suppression capabilities\cite{ref27}. 

These methods have achieved significant progress in perception capabilities. However, they fail to fully utilize the contextual information in images, resulting in a lack of ability to filter out false edges while preserving the edges of small objects.

On the other hand, because of their powerful generalization capabilities, machine learning-based detection methods have dominated edge detection. For instance, EDTER captures contextual features at all scales by using the transformer to separately learn the global and local information of images\cite{ref28}. RankED utilizes the ranking-based losses to enhance the loss function of deep learning, effectively improving the detection accuracy of the algorithm\cite{ref29}. SuperEdge introduces the homography adaptation and dual decoder model into the field of edge detection for the strong performance in terms of generalization\cite{ref30}. EdgeNet integrates features extracted from original images to improve the adversarial robustness of pretrained DNNs\cite{ref31}. DiffusionEdge uses diffusion probabilistic mode to directly generate accurate and clear edge maps\cite{ref32}.

Despite the impressive performance of machine learning models in image edge detection, the generated edge maps are too thick for downstream tasks and tend to miss small-scale edge features. Additionally, the high training costs of these models hinder miniaturization. 

Therefore, we believe that an edge detector that can accurately perceive edges at all scales and filter out false edges without pre-training is necessary. 

\subsection{Discrete-time State Space Model}
The latest success in the field of deep learning, the Mamba architecture, has shown performance comparable to or surpass that of Transformers, especially in context-sensitive domains\cite{ref19}–\hspace{-0.005em}\cite{ref21}. The core of the Mamba architecture is a deep learning network based on the discrete-time state space model (DT-SSM). Its training parameters include matrices A, B, C, and D, and various weights \cite{ref18}\cite{ref33}. The DT-SSM is referred to
\begin{equation}
\left\{\begin{matrix}
x_{k}=Ax_{k-1}+Bu_{k}\\[1mm]
y_{k}=Cx_{k}+Du_{k},\\[1mm]
\end{matrix}\right.
\label{equ1}
\end{equation}
The first equation is the state equation, which updates the state variables of the model based on the input data. The second equation is the output equation, which produces the output based on the model's input and state. Specifically, $u_{k}$ represents the input data to the model, $y_{k}$ is the model's output result, and $x_{k}$ is the model's state variable. $A$ is the system matrix, $B$ is the input matrix, $C$ is the output matrix, and $D$ is the direct transmission matrix. The powerful contextual memory capabilities and simple structure, are exactly what we need. This indicates that the state space model holds significant potential for edge detection tasks.

\subsection{Accelerating Strategy for Edge Detection}
As a fundamental task, the execution speed and efficiency of edge detection are crucial metrics. Although some studies have considered this and improved the processing speed of algorithms \cite{ref13}, most research has not considered this aspect.

Overall, edge detection acceleration tasks can be divided into hardware-level strategies and software-level strategies. Compared to software-level strategies, hardware-level strategies often achieve better performance and lower power consumption because they allow for optimization at the circuit level\cite{Hardware_accelerator_1, Hardware_accelerator_2}. For instance, J. Lee et al. \cite{Hardware_accelerator_3} utilize field-programmable gate arrays (FPGAs) to accelerate the Canny algorithm, achieving 48\% of area and 73\% execution time savings. J. Tian et al. \cite{Hardware_accelerator_4} implemente fast edge detection with memristive operator, which achieved a 50\% reduction in processing time and demonstrated  robust performance in noisy image processing. time.

Memristors are important electronic components in the field of hardware acceleration\cite{ref34}\cite{ref35} as the high efficiency, low power consumption, and small size of memristors make itself a significant research focus in the fields of electronic engineering, artificial intelligence, and neuroscience \cite{ref36}–\hspace{-0.005em}\cite{ref38}. Therefore, we believe it is essential to utilize memristors to accelerate EDCSSM, effectively enhance execution efficiency of algorithm and provide practical value.

\section{The State Space Model for Edge Detection}
This section introduces the architecture of the State Space Model for Edge Detection (EDCSSM) algorithm. The complete architecture of the EDCSSM algorithm is shown in Fig. \ref{fig2}. EDCSSM consists of two parts: an edge detection module called SAIM and a complementary post-processing structure. 

SAIM can be further divided into two modules: the Empirical Cognition Operator and the Intuitive Cognition Operator, corresponding to the state equation and the output equation of the state space model, respectively. The mathematical representation of SAIM is represented by:
\begin{equation}
\left\{\begin{matrix}
 \overline{x}^{n+2}_{k}=a{\times}A^{n}{\ast}p(x^{n+2}_{k-1})+b{\times}B^{n}{\ast}p(u^{2n+1}_{k})\\[1mm]
y_{k}=f(C^{n}{\ast}\overline{x}^{n+2}_{k})+D^{n}{\ast}u^{n}_{k},\\[1mm]
x^{n+2}_{k}=c{\times}\overline{x}^{n+2}_{k}+d{\times}D^{n}{\ast}p(u^{2n+1}_{k}),\\[1mm]
\end{matrix}\right.
\label{equ2}
\end{equation}

where $a$, $b$, $c$, $d$ represent weighting factors, while $A^{n}$, $B^{n}$, $C^{n}$, $D^{n}$ denote convolution kernels with dimensions $n{\times}n$. 
$x^{n}_{k}$ represents the $n{\times}n$-sized state variable obtained at the k-th time step. $u^{n}_{k}$ represents the $n{\times}n$-sized image text input at the $k$-th time step. $f(\cdot)$ denotes the center pixel sampling, and $p(\cdot)$ represents zero-padding around the variable to ensure consistent dimensions. ``${\ast}$'' represents convolution operation. ``$\times$'' denotes element-wise multiplication of a constant with each element of the matrix.

Specifically, the values of these parameters are given in Equation  (\ref{equ3:all}). Here, $a, b, c, d$ is determined by the experimental results of the algorithm, while the remaining parameters are fixed. 
\begin{subequations}\label{equ3:all}
\begin{align}
A_x &= \begin{bmatrix}
-1 & -0.5 & 0 \\
-0.5 & 0 & -0.5 \\
0 & -0.5 & -1
\end{bmatrix}, \quad 
A_y = A_x^T \label{eq:1} \\
B_x &= \begin{bmatrix}
v - v^2 & 2v & -1 \\
v - v^2 & 2v & -2 \\
v - v^2 & 2v & -1
\end{bmatrix}, \quad
B_y = B_x^T, \quad v = 1.3 \label{equ3:2} \\
C_x &= D_x, \quad C_y = D_y \label{equ3:3} \\
D_x &= \begin{bmatrix}
-1 & 0 & 1 \\
-2 & 0 & 2 \\
-1 & 0 & 1
\end{bmatrix}, \quad
D_y = D_x^T \label{eq:4} \\
a &= 0.8, \quad b = 1, \quad c = 0.8 \quad d = 1\label{equ3:5}
\end{align}
\end{subequations}

Ignoring the superscripts, weighting factors, and $f(\cdot)$, $p(\cdot)$ functions, Equation (\ref{equ2}) can be simplified as:
\begin{equation}
\left\{\begin{matrix}
 \overline{x}_{k}=A{\ast}x_{k-1}+B{\ast}u_{k}\\[1mm]
y_{k}=C{\ast}\overline{x}_{k}+D{\ast}u_{k},\\[1mm]
x_{k}=\overline{x}_{k}+D{\ast}u_{k},\\[1mm]
\end{matrix}\right.
\label{equ4}
\end{equation}
The simplified result is similar to Equation (\ref{equ1}). The difference lies in the introduction of low-dimensional information from the direct transmission matrix into the state variables, which is represented by the arrow $X_{LK}$ from the Intuitive Cognition Operator to the Empirical Cognition Operator in Fig.\ref{fig2}. This approach ensures that SAIM can learn information at all scales. 

According to Fig. \ref{fig2}, the post-processing part mainly consists of four major components: gradient computation, non-maximum suppression, double threshold detection, and wind erosion. 
\begin{enumerate}
    \item Gradient computation includes calculating both the gradient magnitude and the gradient direction. Given the horizontal and vertical gradient values \( G_x(x, y) \) and \( G_y(x, y) \) output by EDCSSM, the gradient magnitude is:
\begin{equation}
G(x, y) = \sqrt{G_x(x, y)^2 + G_y(x, y)^2},\\[1mm]
\label{equ5}
\end{equation}

The gradient direction is:
\begin{equation}
  \theta(x,y) = \arctan\left( \frac{G_x(x,y)}{G_y(x,y)} \right),\\[1mm]
\label{equ6}
\end{equation}
	 \item Non-maximum suppression refines edges by preserving local maxima. It determines two neighboring pixels based on the pixel's gradient direction \(\theta(x, y)\). If the gradient magnitude of the current pixel is not the largest compared to its two neighbors, it is set to zero. 
 	\item Hysteresis thresholding is crucial for generating pixel-level edges. Specifically, it sets the pixel intensity to 255 if it is greater than the high threshold, and sets the pixel intensity to 0 if it is less than the low threshold. For pixel intensities between the two thresholds, the pixel value is set to 50. Then, it checks the surrounding pixels of those with a value of 50. If any of the surrounding pixels have an intensity of 255, the pixel value is set to 255, otherwise, it is set to 0.
	\item The wind erosion algorithm filters out false edges from the edges while preserving true edges that are significant to human visual perception. The specific pseudocode is as follows:
\begin{algorithm}
\caption{Wind Erosion}
\KwIn{The binary edge map $E_i$, which is produced by the hysteresis thresholding method}
\KwOut{Processed edge map $E_{\text{out}}$}

\Begin{
1. \textbf{Find Boundarys:}Find edges adjacent to blank areas and cut off branches extending inward. The results are categorized into the set $CE$\\
2. \textbf{Process Long Edges} Calculate the mean length $E_{mean}$ of all edges and split edges longer than  $2{\times}E_{mean}$ at their junctions. Then, recalculate the mean length $E_{mean}$ of all edges and classify edges longer than $p_{mean}{\times}E_{mean}$ into the set PT \\
\hspace*{2em}\tcp{
{\fontsize{8}{12}\selectfont
$p_{mean}$ is a designed value
} 
}
3. \textbf{Split Edges:} Split all edges at their junctions and record the corresponding parent and child edges for each split edge \\
4. \textbf{Clear Edges:} Remove spurs from the edges and delete edges shorter than $L_{t}$ \\
\hspace*{2em}\tcp{
{\fontsize{8}{12}\selectfont
$L_{t}$ is a predefined threshold
}
} 
5. \textbf{Restore Junctions:} Assume a parent edge has a total of $C_{a}$ child edges. Restore the parent edge according to the following rules: if the number of deleted child edges $C_{cut}$ meets both of the following conditions: 1. $C_{cut}<C_{t}$ 2. $C_{cut}<C_{a}{\times}p_{t}$, then restore the parent edge and the remaining child edges. Otherwise, delete the parent edge and corresponding child edges \\
\hspace*{2em}\tcp{   
{\fontsize{8}{12}\selectfont
$C_{t}$ is a predefined threshold, $p_{t}$ is a predefined ratio that falls within the range (0,1)
}
} 
6. \textbf{Restore Protected Edges:} Restore the edges in the set PT and filter out the short edges \\
7. \textbf{Restore Boundarys:} Restore the edges in the set CE \\
\Return The restored edge map $E_{\text{out}}$
}
\end{algorithm}
\end{enumerate}
\section{Circuits Accelerator for EDCSSM}
SAIM is the most computationally intensive part of EDCSSM. To enhance the processing speed and computational efficiency of EDCSSM, we design a parallel analog computing circuit based on a memristor crossbar array to perform the SAIM computation tasks. 

\begin{figure}[htbp]
    \centering
    \begin{subfigure}[b]{0.46\textwidth}
        \centering
        \includegraphics[width=\textwidth]{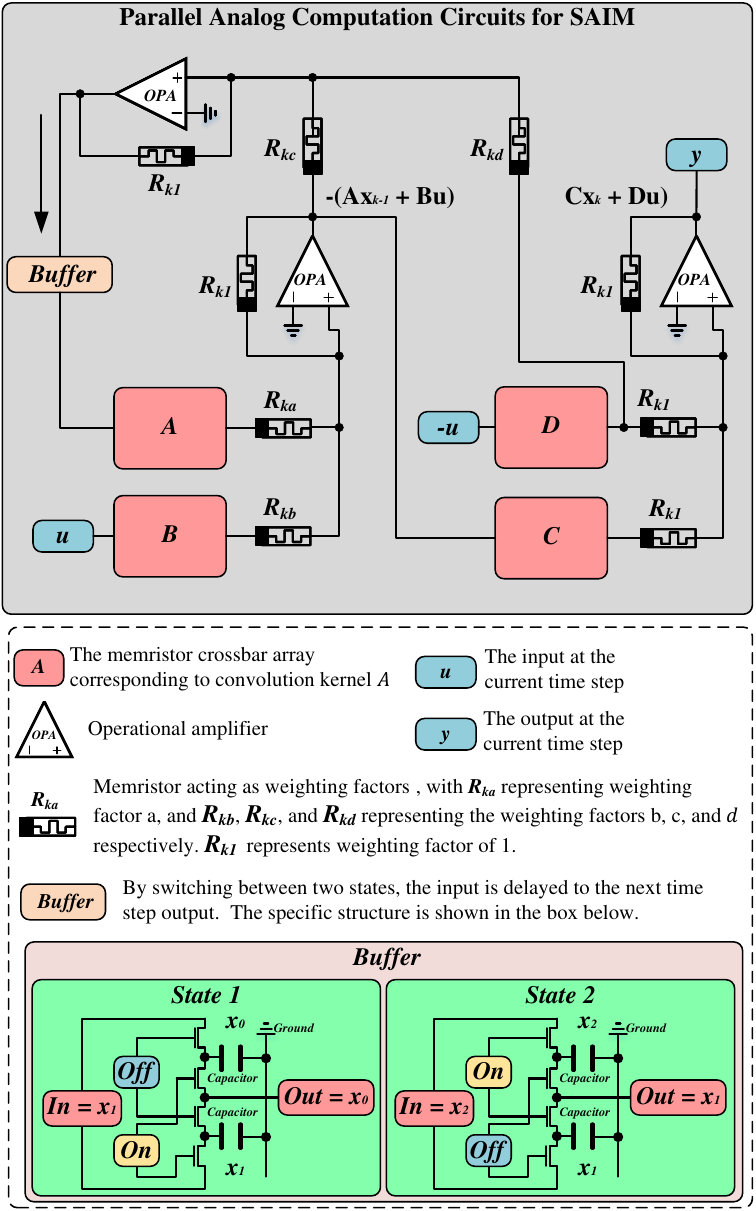}
    \end{subfigure}
  \caption{
Parallel analog computation circuits for SAIM} 

\label{fig3}
\end{figure}
\begin{figure}[htbp]
    \centering
    \begin{subfigure}[b]{0.46\textwidth}
        \centering
        \includegraphics[width=\textwidth]{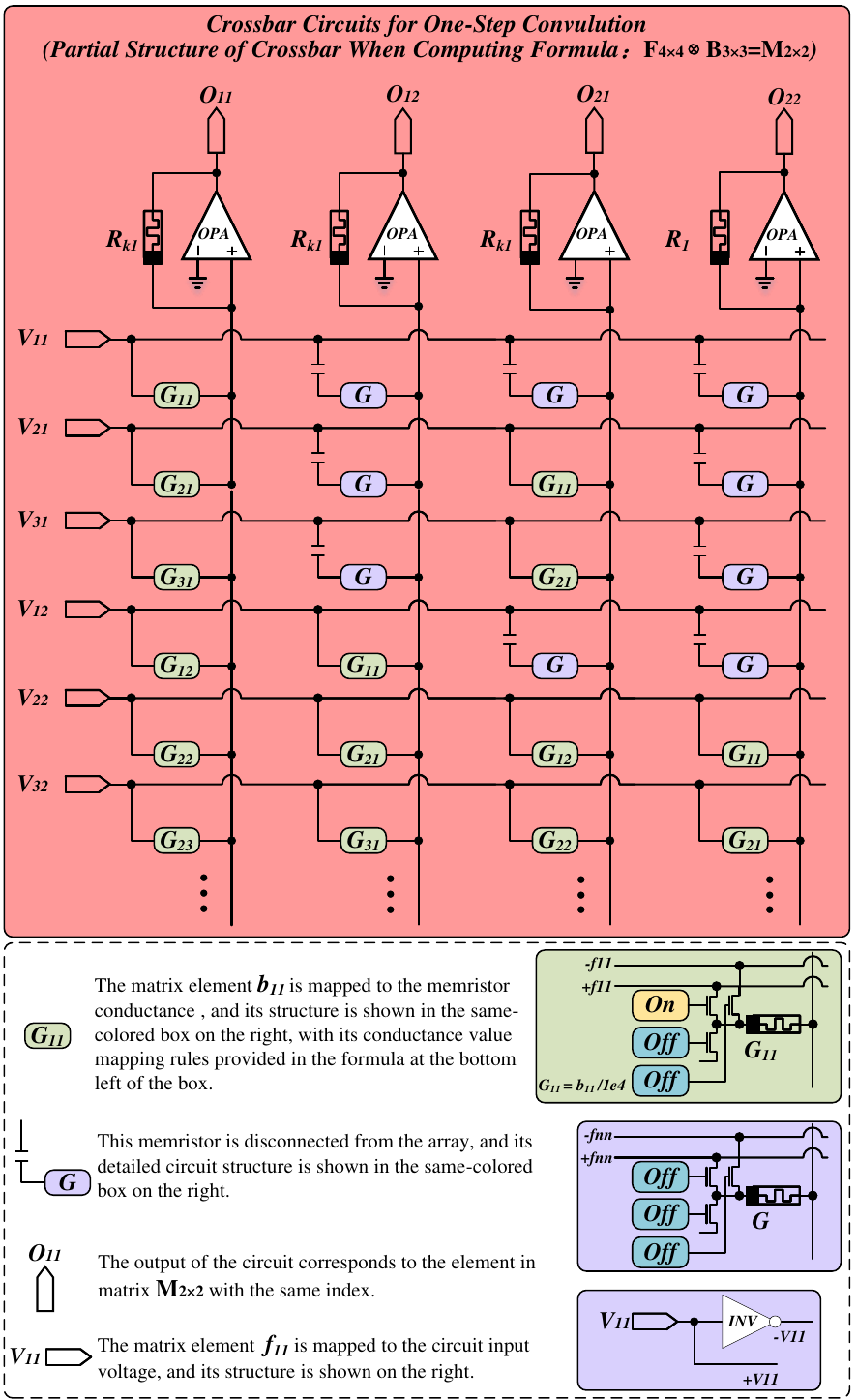}
    \end{subfigure}
  \caption{
Crossbar circuits for one-step convulution} 

\label{fig4}
\end{figure}

The structure of the circuit accelerator is shown in Fig. \ref{fig3}. The input to the accelerator is the image text tensors, as detailed in Equation (\ref{equ2}).  The memristor crossbar array is the core module of the circuit accelerator. It performs convolution tasks through parallel analog computation. The details are illustrated in Fig. \ref{fig4}.  It is important to note that the size of the crossbar array is scalable. The size of the crossbar is determined by both the size of the convolution kernel and the size of the input tensor.  

To illustrate the computational principles of the circuit, we first consider the following convolution operation:
\begin{equation}
Y_{n-m+1}=X_{n}{\ast}K_{m},\\[1mm]
\label{equ7}
\end{equation}
Here, $X$ is an image of size $n{\times}n$, $K$ is a convolution kernel of size $m{\times}m$, and $Y$ is the result of size ${(n-m+1)}{\times}{(n-m+1)}$. For the crossbar, the relationship between the input and output voltages of the crossbar array can be derived from the circuit topology and Kirchhoff's laws, which is specially described as:  
\begin{equation}
O_{ij}=-R_{k1}\sum_{u=1}^{n}\sum_{v=1}^{n}{G_{u,v}}\times{V_{(i+u-1,j+v-1)}},\\[1mm]
\label{equ8}
\end{equation}
Here, $O_{ij}$ is the output voltage value at the crossbar,  $G_{u,v}$ is the conductance value of the memristor, $R_{k1}$  is the resistance value of the memristor in the feedback loop of the operational amplifier,  $V_{(i+u-1,j+v-1)}$ is the input voltage value of the crossbar. The mapping rules from algorithm to circuit are crucial for linking Equation (\ref{equ7}) and Equation (\ref{equ8}). In this paper, the mapping rules are described as Table \ref{tab1}. 
\begin{table}[htbp]
\begin{center}
\begin{threeparttable}
\caption{Parameter Relationships between Algorithm and Circuits}
\scriptsize
\begin{tabular}{ccc}
\toprule
 \textbf{\textit{\rm Algorithm Parameters}}& \textbf{\textit{\rm Circuits Parameters}}&\textbf{\textit{\rm Mapping Relationships}}\\
\midrule
 & & \\[-2.5mm]
  $x_{i,j}$ & $V_{i,j}$  &$ V_{i,j}$=$P_{V}\times x_{i,j} \mathrm{(V)}$ \\[0.5mm]
  $k_{u,v}$ & $G_{u,v}$  &  $G_{u,v}=P_{G}\times k_{_{u,v}}\mathrm{(S)}$ \\[1mm]
$y_{i,j}$ & $O_{i,j}$ & $-O_{i,j}/R_{k1}=P_{V}{\times}P_{G}{\times}y_{i,j}$\\[0.5mm]
\bottomrule
\multicolumn{3}{l}{}
\label{tab1}
\end{tabular}
 \begin{tablenotes}
        \footnotesize
        \item[*]  $x_{i,j}$: Grayscale value of the pixel at coordinates $(i,j)$ in image $X_{n}$.
		  \item[*]  $k_{u,v}$: Value of the element at coordinates $(u,v)$ in the convolution kernel $K_{m}$.
		  \item[*]  $y_{i,j}$: Grayscale value of the pixel at coordinates $(i,j)$ in the convolution result $Y_{n-m+1}$.
		  \item[*]  $V_{i,j}$: Input voltage value corresponding to $x_{i,j}$.
		  \item[*]  $G_{u,v}$: Conductance value of the memristor corresponding to $k_{u,v}$.
		  \item[*]  $O_{i,j}$: Output voltage value of the crossbar corresponding to $y_{i,j}$.
		  \item[*]  $P_{V}$: A predefined scaling factor, where $P_{V}=10^{-2}$ in this paper.
		  \item[*]  $P_{G}$: A predefined scaling factor, where $P_{G}=10^{-4}$ in this paper.
		  \item[*]  $R_{k1}$: Resistance value of the memristor in the feedback loop of the operational amplifier
 \end{tablenotes}
	\end{threeparttable}
\end{center}
\end{table}

Based on Table \ref{tab1} and Equations (\ref{equ7}) and (\ref{equ8}), the relationship between the algorithm and the circuit can be derived as: 
\begin{equation}
O_{n-m+1}=-R_{k1}{\times}{10^{-8}}{\times}Y_{n-m+1},\\[1mm]
\label{equ9}
\end{equation}
The above equation indicates that $Y_{n-m+1}$ and $O_{n-m+1}$ differ only by a coefficient, which can be eliminated during the  Analog-to-Digital (AD) conversion  process. Therefore, the memristor crossbar array can be used to achieve fast convolution computation.

\section{Experiments on Edge Detection}
In this section  we test the performance of the algorithm and compare it with state-of-the-art methods.
\subsection{Experiments Datasets}
We conduct experiments on three edge detection datasets: BSDS \cite{ref_dataset_BSDS500}, BIPED \cite{ref_dataset_BIPED}, NYUD \cite{ref_dataset_NYUD} and PASCAL \cite{ref_dataset_PASCAL}.  BSDS500 contains 200, 100, and 200 images in the training, validation, and test set, respectively. Each image has 4-9 annotators to determine the final edge ground truth. BIPED contains 250 annotated images of outdoor sceness, divided into a training set comprising 200 images and a testing set containing 50 image. All images are carefully annotated at single-pixel width by experts in the computer vision field. For the PASCAL and NYUD datasets, we do not conduct direct testing. Instead, we selecte images with different characteristics from BSDS500, PASCAL, and NYUD to create a small dataset for ablation study.
\subsection{Performance Metrics}
On one hand, the introduction of the wind erosion algorithm causes jagged fluctuations in the PR curve, indicating that conventional metrics are not suitable for our work. On the other hand, since our algorithm integrates post-processing steps, the prediction edge thickness is generally 1-2 pixels.  Using a one-to-one matching method to compare our algorithm's edges with manually labeled edges would lead to significant errors due to slight positional discrepancies. Moreover, we want to take edge thickness into account to avoid localization errors caused by thick edges. Therefore, we designe a specialized evaluator that considers these issues when calculating TP, FP, and FN. Based on this, we compute the algorithm's Average Contour Length (ACL), Average Structural Similarity Index (SSIM), Optimal Dataset Scale (ODS), Optimal Image Scale (OIS), and Area Coverage (AC) on the corresponding datasets. The specific methods for calculating TP, FP, and FN are as follows:
\begin{algorithm}
\caption{Modified  Metrics}
\KwIn{The binary edge map $normal$, which is output by the algorithm, the Ground Truth $GT$}
\KwOut{True Positive $TP$, False Positive $FP$, False Negative $TN$}

\Begin{
    \textbf{1. Initialize Counters:} \\
    Initialize $TP$, $FP$, $FN$ to 0.
    
    \textbf{2. Iterate Over Each Pixel:} \\
    \For{each pixel $(i, j)$ in $normal$}{
        \eIf{$GT[i, j]$ is an edge pixel}{
            Count edge pixels in the $5 \times 5$ neighborhood of $(i, j)$ in $normal$.\\
            \eIf{edge pixel count is between 3 and 12}{
                $TP$ +=1.
            }{
                $FN$ +=1.
            }
        }{
            Count edge pixels in the $5 \times 5$ neighborhood of $(i, j)$ in $normal$.\\
            \If{edge pixel count is greater than or equal to 12}{         
			    $FP$ +=1. 
            }
        }
    }
    
    \textbf{4. Return True Positive, False Positive and False Negative:} \\
    \Return $TP$, $FP$, $FN$ 
}
\end{algorithm}

\subsection{Implementation Details}
EDCSSM does not require pre-training and can be implemented directly using OpenCV. However, the algorithm's weighting parameters $a$, $b$, $c$ and $d$ need to be determined (see Equation (\ref{equ3:5})). Specifically, the upper and lower thresholds are calculated using a fixed method. For each image, the weights $a$, $b$, $c$ and $d$ are increased from 0 to 2 in steps of 0.1.  Based on this, the optimal F-measure for each image is calculated, and the corresponding weights are recorded. The mode of these recorded weights is taken as the final weight.  Then,   the relationship between the algorithm's thresholds and its performance is studied on the BIPED dataset. The upper threshold $H_{threshold}$ is increased from 0 to 255 in steps of 2.55. The lower threshold $L_{threshold}$ is set to $L_{threshold}=0.95{\times}H_{threshold}$. Finally, we conduct ablation studys on the combined dataset. All studys are conducted on a platform equipped with an I9-12900H CPU and 32GB RAM, without using GPU.

\subsection{Comparison with State-of-the-art}
\begin{figure*}[htbp]
  \centering
  \includegraphics[width=\textwidth]{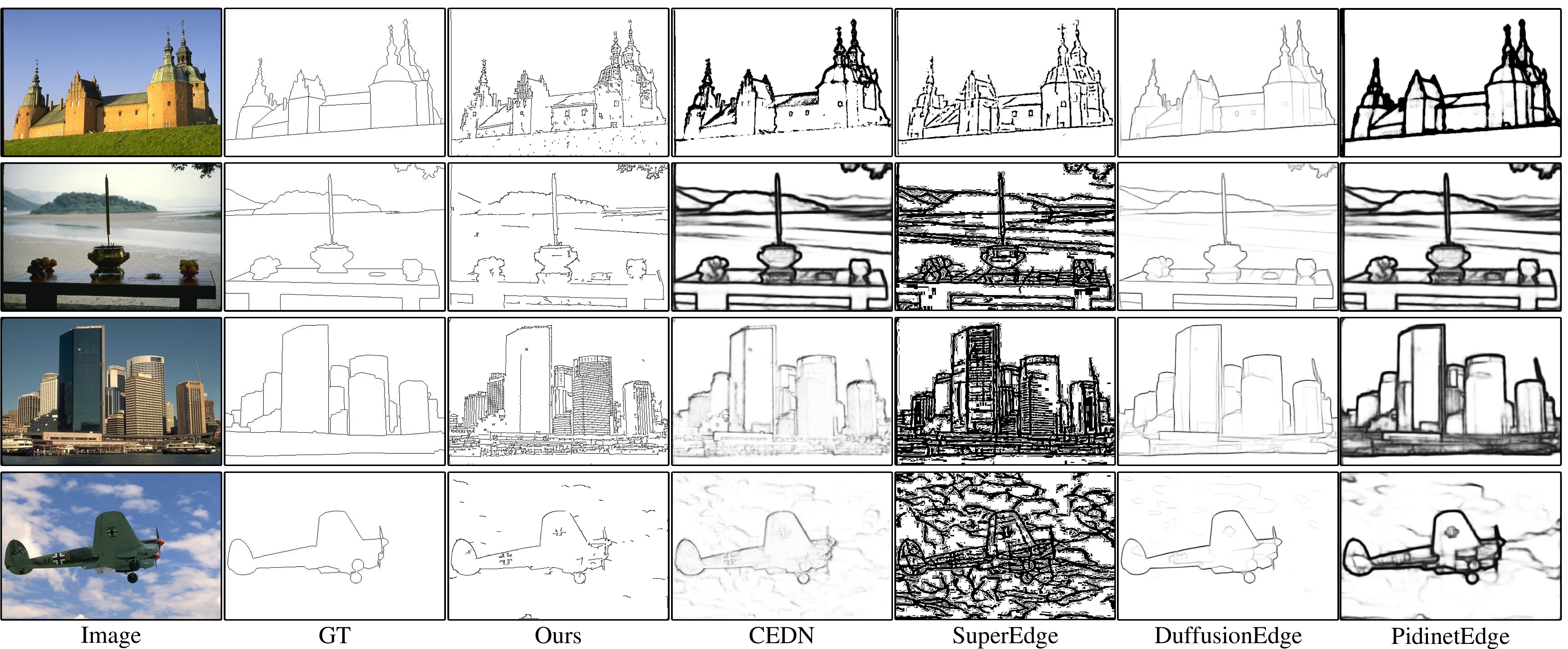}
  \caption{Performance Comparisons on BSDS500 dataset with previous state-of-the-arts.}
  \label{fig5}
\end{figure*}
We compare our model (without wind erosion) with previous works, including Canny\cite{ref7}, HED\cite{ref12}, CEDN\cite{ref13}, PiDiNet\cite{ref_SOTA_PIDINET}, SuperEdge\cite{ref_SOTA_SUPERedge}, and DiffusionEdge et al.\cite{ref32}. The best results of all methods are either taken from their publications (ODS, OIS, AC) or tested using the provided implementations on the corresponding datasets (ACL, SSIM). 
\begin{table}[htbp]
\caption{Performance Comparisons on BSDS500}
\begin{center}
\resizebox{.48\textwidth}{!}{
\begin{threeparttable}
\scriptsize
\begin{tabular}{c|cccccc}
\bottomrule
\textbf{\textit{\rm Methods}} & \textbf{\textit{\rm Pretraining}}& \textbf{\textit{\rm ACL}}&\textbf{\textit{\rm SSIM}}&\textbf{\textit{\rm ODS}} & \textbf{\textit{\rm OIS}} & \textbf{\textit{\rm AC}}\\
\hline
& & & & &\\[-2mm]
Canny &  $\times$ & 30.3800 & 0.9751 & - & - & -\\[0.75mm]
CEDN &  $\checkmark$  & 22.4241 &  0.9861 & - & - & -\\[0.75mm]
RCF &  $\checkmark$  & 23.3739 & 0.9189 & 0.585 & 0.604 & 0.189\\[0.75mm]
SuperEdge &  $\checkmark$ & - & - & 0.672 & 0.686 & -\\[0.75mm]
HED & $\checkmark$  & 9.0102 & 0.9399 & 0.588 & 0.608 & 0.215\\[0.75mm]
PidinetEdge & $\checkmark$ & 59.3883 &  0.9249  & 0.578 & 0.587 & 0.202\\[0.75mm]
EDTER & $\checkmark$ & 33.8938 &  0.9472 & 0.698 & 0.706 & 0.288\\[0.75mm]
DuffusionEdge & $\checkmark$ & 8.7765 & 0.9948 & 0.749 & 0.754 & 0.476\\[0.75mm]
Ours & $\times$ & \textbf{23.9149} & \textbf{0.9938} & \textbf{0.6024} & \textbf{0.6164} & \textbf{0.6034}\\[0.75mm]
\toprule

\end{tabular}
	\end{threeparttable}
\label{tab2}
}
\end{center}
\end{table}

\textbf{On BSDS}. Based on Table \ref{tab2} and Fig. \ref{fig5}, several conclusions can be drawn: (a) The proposed algorithm achieves the best AC value. Generally, a higher AC value indicates that the prediction edges that match the ground truth are thinner. This demonstrates that the edges captured by the algorithm are very precise, completely avoiding the issue of decreased localization accuracy caused by thick edges. 
(b) The algorithm's ODS and OIS values are not high. This is because our algorithm detects potential edges in the image by learning full-scale image texture features, which means it perceives more comprehensive edge features than those manually labeled (such as the textures on building facades, insignias on airplanes, and window frames on ships, as shown in Fig. \ref{fig5}). These additional features lead to the lower ODS and OIS values. 
(c) The algorithm achieves a high ACL value and an SSIM value, indicating that the prediction edges are highly continuous and structurally similar to the ground truth. In comparison, PiDiNet and EDTER can achieve highly continuous edges but have lower structural similarity, whereas the DiffusionEdge algorithm achieves good structural similarity but lacks edge continuity.
 
\textbf{On BIPED}. We further test EDCSSM on the BIPED dataset and compare it with algorithms including CEDN, HED, EDTER, PiDiNet, and DiffusionEdge. The results are shown in Table \ref{tab3}. EDCSSM continued to achieve very precise edge prediction results, specifically with an AC value second only to DiffusionEdge. Additionally, it is worth noting that our method performed better in terms of ODS and OIS on the BIPED dataset compared to BSDS500, as the finer manually labeled edge features in BIPED. This further demonstrates the full-scale edge perception capability of our method. These characteristics are also reflected in the higher SSIM and ACL values. 
\begin{table}[htbp]
\caption{Performance Comparisons on BIPED}
\begin{center}
\resizebox{.48\textwidth}{!}{
\begin{threeparttable}
\scriptsize
\begin{tabular}{c|cccccc}
\bottomrule
\textbf{\textit{\rm Methods}} & \textbf{\textit{\rm Pretraining}}& \textbf{\textit{\rm ACL}}&\textbf{\textit{\rm SSIM}}&\textbf{\textit{\rm ODS}} & \textbf{\textit{\rm OIS}} & \textbf{\textit{\rm AC}}\\
\hline
& & & & &\\[-2mm]
CEDN &  $\checkmark$  & 12.8573 &  0.9935 & - & - & -\\[0.75mm]
HED &  $\checkmark$  & 12.1613 & 0.9448 & 0.387 & 0.404 & -\\[0.75mm]
DexiNed & $\checkmark$  & - & - & 0.859 & 0.867 & 0.295\\[0.75mm]
PidinetEdge & $\checkmark$ & - & - &  0.868  & 0.876 & 0.232 \\[0.75mm]
EDTER & $\checkmark$ & - &  - & 0.893 & 0.898 & 0.26 \\[0.75mm]
DuffusionEdge & $\checkmark$ & 30.5383 & 0.9965 & 0.899 & 0.901 & 0.849\\[0.75mm]
Ours & $\times$ & \textbf{27.4516} & \textbf{0.9975} & \textbf{0.7931} & \textbf{0.8036} & \textbf{0.8453}\\[0.75mm]
\toprule

\end{tabular}
	\end{threeparttable}
\label{tab3}
}
\end{center}
\end{table}
Overall, EDCSSM demonstrates excellent full-scale edge perception and good noise suppression capabilities while ensuring edge continuity and extracting thin edges.
\subsection{Ablation Study}
Ablation studies are conducted on the combined dataset. The combined dataset consists of 81 different types of images from BSDS500, NYUD, and PASCAL. 

\textbf{The effect of state variable}. We first conduct experiments to verify the impact of the state variable $x_k$. The quantitative results are summarized in Table \ref{tab4}.  Specifically,  $SAIM_{zero}$. denotes the model with state variables removed, where $A= B = C= 0^{(3 \times 3)}$. In this configuration, SAIM loses its inference and learning capabilities, degrading into a fixed convolution operator. 
\begin{table}[htbp]
\caption{Effectiveness of State Variable in EDCSSM.}
\begin{center}

\begin{threeparttable}
\scriptsize
\begin{tabular}{c|ccc}
\bottomrule
\textbf{\textit{\rm Model}} &\textbf{\textit{\rm ODS}} & \textbf{\textit{\rm OIS}} & \textbf{\textit{\rm AC}}\\
\hline
& & &\\[-2mm]
$SAIM_{zero}$ & 0.7948 & 0.8262 & 0.8119 \\[0.75mm]
$SAIM$ & 0.8241 & 0.8574 & 0.8540 \\[0.75mm]
\toprule

\end{tabular}
	\end{threeparttable}
\label{tab4}

\end{center}
\end{table}

It can be observed that the introduction of state variables enhances the algorithm's edge perception capability and effectively refines the edges, making the extracted edges more precise. This is specifically demonstrated by the fact that the metrics for $SAIM$ are superior to those for $SAIM_{zero}$.

\textbf{The effect of flipping operations}. We study the impact of different flipping operations. The results are evaluated using Average Contour Length (ACL) and Average Structural Similarity Index (SSIM). Specifically, ACL reflects the average length of detected edge segments, while SSIM indicates the structural similarity between the results and the ground truth, with values closer to 1 representing higher similarity. The relevant results are shown in Table \ref{tab5}.
\begin{table}[htbp]
\caption{Effectiveness of Flip operation.}
\begin{center}

\begin{threeparttable}
\scriptsize
\begin{tabular}{cc|cc}
\bottomrule
\textbf{\textit{\rm Horizon Flip}} & \textbf{\textit{\rm Vertical Flip}}&\textbf{\textit{\rm ACL}}&\textbf{\textit{\rm SSIM}}\\
\hline
  &    & & \\[-2mm]
$\times$ & $\times$  & 24.2 & 0.9970   \\[0.75mm]
$\checkmark$  &  $\times$  & 28.7 &  0.9972  \\[0.75mm]
$\times$ &  $\checkmark$  & 28.6  & 0.9972   \\[0.75mm]
$\checkmark$ & $\checkmark$  & 28.7 & 0.9972\\[0.75mm]
\toprule

\end{tabular}
	\end{threeparttable}
\label{tab5}

\end{center}
\end{table}

It can be observed that without any flipping operations, the extracted edges are shorter, resulting in more fragmented edge maps. This issue is particularly noticeable in some cases. Implementing any type of flipping operation increases the edge length, making edges more coherent and avoiding fragmented edges. However, the flipping operations have minimal impact on SSIM, as SSIM is primarily influenced by the edge detection algorithm itself rather than the post-processing steps.

\begin{figure}[htbp]
    \centering
    \begin{subfigure}[b]{0.46\textwidth}
        \centering
        \includegraphics[width=\textwidth]{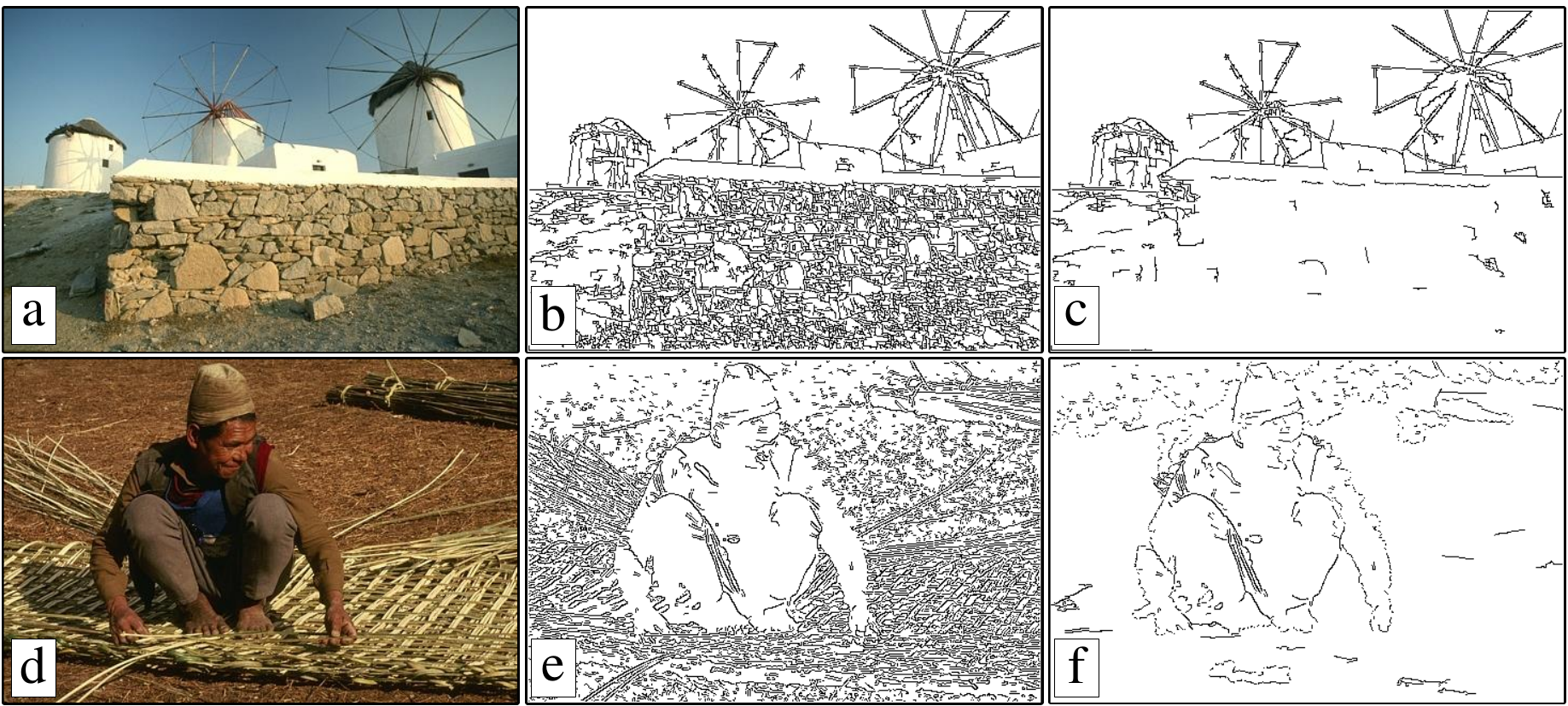}
    \end{subfigure}
  \caption{
\textbf{Impact of wind erosion}. 
The wind erosion algorithm effectively filters out false edges while preserving visually significant edges at all scales. (a, d): Input images from BSDS500. (b, e): The edges obtained with non-optimal thresholds. (c, f): The edges after wind erosion}
\label{fig6}
\end{figure}

\textbf{The effect of wind erosion}. We study the impact of the wind erosion algorithm on the prediction edges. To demonstrate the effectiveness of the algorithm, all edges are obtained without using the optimal threshold. The results are shown in Fig. \ref{fig6}.

According to the results in Fig. \ref{fig6}, the wind erosion algorithm effectively filters out false edges (such as gaps between stones in wall and textures on the woven straw mat) while retaining visually significant edges (such as  the supports of windmill blades and boundaries between people and background). This demonstrates that the erosion algorithm effectively performs the task of filtering and preserving edge maps, thereby proving the algorithm performance.

\section{Simulation of Crossbar Circuits Accelerator}
\begin{figure}[htbp]
    \centering
    \begin{subfigure}[b]{0.46\textwidth}
        \centering
        \includegraphics[width=\textwidth]{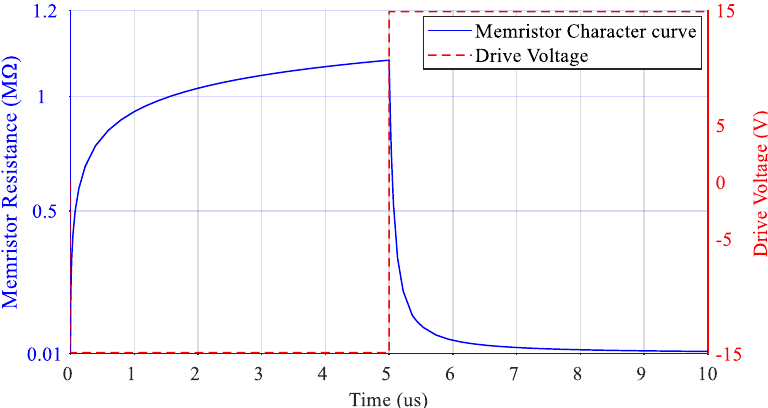}
    \end{subfigure}
  \caption{
Memristor Adjustment Characteristics Curve.} 

\label{fig7}
\end{figure}

The simplified circuit simulation is conducted with PSpice on Candence SPB Release 22.1 version and the memristor model is derived from the relevant work of Li Y et al. \cite{ref39}. The key performance metrics of the operational amplifier model employed in this study are provided in Table \ref{OPAMP}, and the parameters of the memristor model are specified in Table \ref{tab6}. The adjustment characteristics curve of the memristor is obtained, as shown in Fig. \ref{fig7}.
\begin{table}[H]
\caption{Key Performance Metrics of Operational Amplifier }
\begin{center}

\begin{threeparttable}
\scriptsize
\begin{tabular}{c|c}
\bottomrule
 \textbf{\textit{\rm Parameter}} & \textbf{\textit{\rm Value}}\\

\hline
 & \\[-2.5mm]
  Unity Gain Bandwidth  &  $8.36935  \mathrm{(MHz)}$ \\[0.5mm]
  Phase Margin  &  $67.0622 \mathrm{(^\circ)}$ \\[1mm]
  Open-Loop Gain &  $91.344  \mathrm{(dB )}$  \\[0.5mm]
  Slew Rate & $5.905   \mathrm{(V/{\mu}s)}$ \\[0.5mm]
  Average Transition Rate & $3.9313  \mathrm{(V/{\mu}s)}$\\[0.5mm]
  Maximum Step Response Time & $356.1161  \mathrm{(ns)}$ \\[0.5mm]
  Supply Voltage &  $1.8  \mathrm{(V)}$ \\[0.5mm]
  Process Technology &  $180  \mathrm{(nm)}$ \\[0.5mm]
\toprule
\multicolumn{2}{l}{}
\end{tabular}
	\end{threeparttable}
\label{OPAMP}

\end{center}
\end{table}

\begin{table}[htbp]
\caption{Details of Memristor Model Parameters}
\begin{center}

\begin{threeparttable}
\scriptsize
\begin{tabular}{cc|cc}
\bottomrule
 \textbf{\textit{\rm Parameters}}& \textbf{\textit{\rm Values}}& 
 \textbf{\textit{\rm Parameters}}& \textbf{\textit{\rm Values}}\\
\hline
 & & & \\[-2.5mm]
  $V_{on}$(V) & 0.5  & $R_{on}$(M$\Omega$) & $10^{-4}$\\[0.75mm]
	  $V_{off}$(V) & 0.5  &  $R_{off}$(M$\Omega$) &  1.5\\[0.75mm]
  $a_1$, $a_2$ & 100, 20 &   $k_{on}$ & 500 \\[0.75mm]
  $p_1$, $p_2$ & 2.5, 5 & $k_{off}$ & 10 \\[0.75mm]
\toprule
\multicolumn{3}{l}{}
\end{tabular}
\end{threeparttable}
\label{tab6}

\end{center}
\end{table}

\begin{figure*}[htbp]
  \centering
  \includegraphics[width=\textwidth]{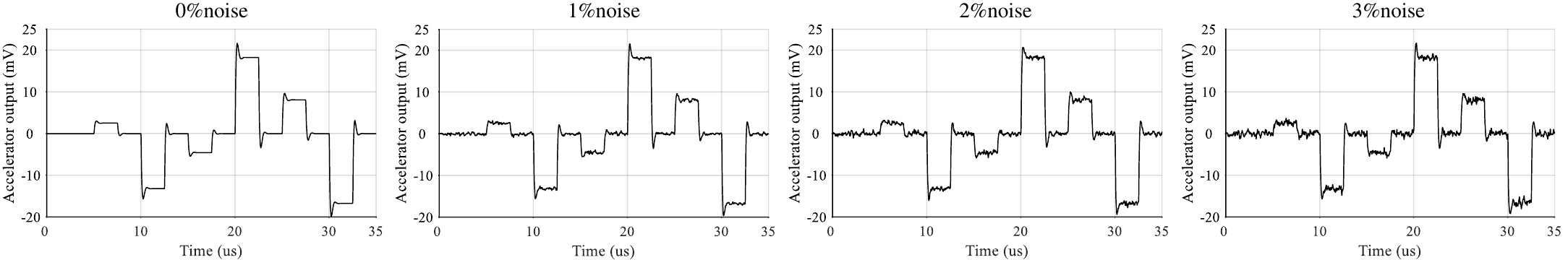}
  \caption{Output Pulse of Accelerator with Different Input Noise.}
  \label{fig8}
\end{figure*}

Based on the characteristic curve, the regulation period of the memristor is 10$\mu$s. The output period of the circuit accelerator is set to 10$\mu$s as well, with an output pulse duty cycle of 0.5. This ensures that there is no interference between consecutive output pulses and improves the accuracy of the accelerator. 

To investigate the anti-interference performance of the circuit, the circuit's performance under different noise interferences is tested. Details are presented in Fig. \ref{fig8} and Table \ref{tab7}. According to Fig. \ref{fig8}, noticeable distortion appears at the rising  edge of the square wave output, regardless of the presence of noise. This distortion is attributed to the step signal from the input circuit causes a step response in the operational amplifier, thereby contaminating the rising edge of the output. To mitigate the impact of distortion on circuit accuracy, sampling should be conducted in the latter half of the square wave. By comparing the output pulse waveforms at different noise levels, it is evident that higher noise levels lead to more severe distortion of the circuit's output signal. It can be predicted that as noise continues to increase, the Analog-to-Digital Converter (ADC) will be unable to distinguish valid outputs from the noise. Therefore, the noise level in practical circuits should be kept low.

\begin{table}[htbp]

\caption{Accuracy of Accelerator without sampling average.}
\begin{center}
\begin{threeparttable}
\scriptsize
\begin{tabular}{ccc}
\bottomrule
 \textbf{\textit{\rm Noise Level }}& \textbf{\textit{\rm Error Value}}& 
 \textbf{\textit{\rm Percentage Error}}\\
(\%)& (mV)& 
 (\%)\\
\midrule
 & & \\[-2.5mm]
  0 & 0.0109 &  0.4367 \\[0.75mm]
  5 & 0.2979 &  11.9151\\[0.75mm]
  10 & 0.4191 &  16.7638 \\[0.75mm]
  20 & 0.5436 &  21.7444 \\[0.75mm]
  30 & 1.1390 &  45.5609 \\[0.75mm]

\toprule
\multicolumn{3}{l}{}
\end{tabular}
\end{threeparttable}
\label{tab7}
\end{center}
\end{table}

Table \ref{tab7} illustrates the maximum output error of circuit when sampling is performed only once in the latter half of the square wave. According to the details, even with only 5\% noise, the maximum output error reached 11\%. It is evident that the impact of noise on circuit output accuracy is significant under this condition. 
\begin{table}[htbp]
\caption{Accuracy of Accelerator with sampling average.}
\begin{center}
\begin{threeparttable}
\scriptsize
\begin{tabular}{ccc}
\bottomrule
 \textbf{\textit{\rm Noise Level }}& \textbf{\textit{\rm Error Value}}& 
 \textbf{\textit{\rm Percentage Error}}\\
(\%)& (mV)& 
 (\%)\\
\midrule
 & & \\[-2.5mm]
  0 & 0.0021 &  0.0855 \\[0.75mm]
  5 & 0.0229 &  0.9158\\[0.75mm]
  10 & 0.0445 &  1.7803 \\[0.75mm]
  20 & 0.0542 & 2.1698\\[0.75mm]
  30 & 0.454 &  1.8144 \\[0.75mm]
\toprule
\multicolumn{3}{l}{}
\end{tabular}
\end{threeparttable}
\label{tab8}
\end{center}
\end{table}
To mitigate the influence of noise, a strategy of averaging after sampling is necessary. Specifically, multiple points should be sampled in the latter half of the square wave, and their average taken as the result. Table \ref{tab8} demonstrates the relationship between circuit accuracy and noise using the sampling average strategy.

Specially, the results in Table \ref{tab8} are obtained by sampling multiple points and averaging them within the interval of 1.2 seconds to 1.4 seconds after the rising edge of the output.  In Table \ref{tab8}, the percentage errors  are all below 2.5\% across different noise levels. Contrasting with Table \ref{tab7}, it is evident that the sampling-averaging strategy effectively mitigates the impact of noise on circuit accuracy, significantly enhancing the circuit's robustness. However, it is important to note that sampling multiple points and computing their average introduces additional computational and time costs. Therefore, efforts should be made to minimize the number of required sampling points.
\begin{table}[htbp]
\caption{Theoretical Time Costs on Images of Different Sizes.}
\begin{center}
\resizebox{.48\textwidth}{!}{
\begin{threeparttable}
\scriptsize
\begin{tabular}{ccc|cc}
\bottomrule
\textbf{\textit{\rm Index}} & \textbf{\textit{\rm Image Resolution}}& \textbf{\textit{\rm Pixel Count}}&\textbf{\textit{\rm Processing Time (s)}} & \textbf{\textit{\rm FPS}}\\

\hline
& & & &\\[-2mm]
1 &  640$\times$480 & 307,200 & 0.0005 & 1887.1\\[0.75mm]
2 &  1280$\times$720  & 921,600 &  0.0014 & 695.6\\[0.75mm]
3 &  1600$\times$900  & 1,440,000 & 0.0024 & 421.1\\[0.75mm]
4 &  1920$\times$1080  & 2,073,600 & 0.0034 & 295.3 \\[0.75mm]
5 & 2560$\times$1440 & 3,686,400 & 0.0063 & 159.3\\[0.75mm]
6 & 2560$\times$2048 &5,242,880 &  0.0116 & 86.5\\[0.75mm]
7 & 3840$\times$2160 & 8,294,400 &  0.0177 & 56.4\\[0.75mm]
8 & 4068$\times$3072& 12,496,896 & 0.0290 & 34.4\\[0.75mm]
9 & 5120$\times$2880& 14,745,600 & 0.0314& 31.9\\[0.75mm]
10 & 6524$\times$4353& 28,398,972 & 0.0587 & 17.0\\[0.75mm]
11 & 8000$\times$4500& 36,000,000 & 0.0746 & 13.4\\[0.75mm]
12 & 9000$\times$4651& 41,859,000 & 0.0901 & 11.1\\[0.75mm]
13 & 9396$\times$5960& 56,000,160 & 0.1181 & 8.5\\[0.75mm]
14 & 9376$\times$6336& 59,406,336 & 0.1225 & 8.2\\[0.75mm]
15 & 10922$\times$6000& 65,532,000 & 0.1376 & 7.3\\[0.75mm]
16 & 11245$\times$6604& 74,261,980 & 0.1722 & 5.8	\\[0.75mm]
17 & 12000$\times$7300& 87,600,000 & 0.1835& 5.4\\[0.75mm]
18 & 10000$\times$10000& 100,000,000 & 0.22907& 4.5\\[0.75mm]
19 & 19944$\times$6309& 125,826,696 & 0.2932& 3.4\\[0.75mm]
20 & 16877$\times$13107& 221,206,839 & 0.5225& 1.9\\[0.75mm]

\toprule
\multicolumn{3}{l}{}
\end{tabular}
	\end{threeparttable}
\label{tab9}
}

\end{center}
\end{table}

	Finally, we accelerate the EDCSSM algorithm using a crossbar array accelerator and investigate the processing speeds of SAIM on images of different sizes. Here, we employed 2000 crossbar array, with each crossbar array containing 4000 memristors. The detailed results are presented in Table \ref{tab9}.

All images have practical significance and are completely independent of each other at different scales. Additionally, the processing time for each image scale is calculated repeatedly, with the best value taken as the final result. 

According to the results in Table \ref{tab9}, after acceleration using the memristor crossbar array, the algorithm maintains a processing speed of 30 FPS at 5K resolution (corresponding to index 9). At 9K resolution (index 12), the processing speed approaches 11 FPS, and at the highest resolution we tested (index 20), the speed remains above 1 FPS. This demonstrates that accelerating the algorithm through parallel-analog computing with the memristor crossbar array effectively enhances the processing speed and provides a feasible method for the practical implementation of the algorithm.

\section{Conclusion}
In this paper, we introduce the first full-scale edge detection algorithm based on a discrete state space model. Through the design of several novel techniques, including a self-learning module and a wind erosion post-processing algorithm, EDCSSM achieves full-scale edge detection and precise localization capabilities comparable to manual feature operators, while also retaining flase edge suppression capabilities akin to deep learning algorithms. This feature is absent in previous algorithms. Additionally, a corresponding architecture-level parallel-analog computing circuit accelerator is designed to complete the core computation tasks of EDCSSM within 5$\mu$s. With the accelerator, the algorithm achieves processing speeds of 30 FPS on 5K images, 11 FPS on 9K images, and approximately 2 FPS on 20K images. This significantly enhances the algorithm's processing speed and efficiency, adding practical value.

	\textbf{Limitations.} EDCSSM extracts precise and accurate edge maps while effectively suppressing most noise by combining with wind erosion. However, its noise suppression in complex scenes is suboptimal, and the post-processing structure is complex. Introducing multiple state variables in the state-space model is a potential improvement direction to enhance learning ability and simplify the post-processing structure. Additionally, the wind erosion algorithm involves many intricate parameters, making self-tuning of these parameters  crucial.

\vspace{-40pt}
\begin{IEEEbiography}[{\includegraphics[width=1in,height=1.25in,clip,keepaspectratio]{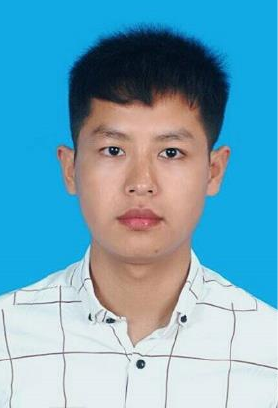}}]{Qinghui Hong}
received the B.S degree and M.S degree in electronic science and technology from Xiangtan University, Xiangtan, China, in 2012 and 2015, respectively, and the Ph.D degree in computer system architecture from Huazhong University of Science and Technology, Wuhan, China, in 2019. He is currently an associate professor with college of computer science and electronic engineering, Hunan University, Changsha 410082, China. He has presided over published more than30 SCI papers, including IEEE TNNLS, IEEE TBCS, IEEE TCAD, IEEE TVLSI, IEEE TCAS-I, etc. His current research interests include memristive neural network, brain-like computing circuits and its application to Artificial Intelligence.
\end{IEEEbiography}
\vspace{-41pt}

\begin{IEEEbiography}[{\includegraphics[width=1in,height=1.25in,clip,keepaspectratio]{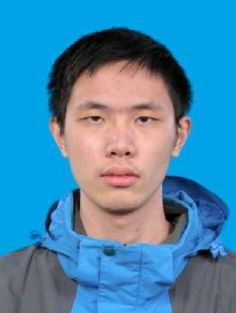}}]{Haoyou Jiang}
 received the B.S. degree in automation from Northeastern University at Qinhuangdao, Qinhuangdao, China, in 2022. He is currently pursuing the M.S. degree with the college of computer science and electronic engineering, Hunan University, Changsha, China. His current research interests are the circuit design of memristors and its application.
\end{IEEEbiography}
\vspace{-41pt}

\begin{IEEEbiography}[{\includegraphics[width=1in,height=1.25in,clip,keepaspectratio]{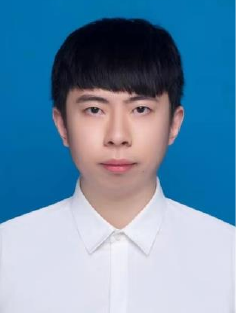}}]{Pingdan Xiao }

received the B.S. degree from Tianjin Sino-German University of Applied Sciences, Tianjin, China, in 2021. He is currently pursuing the M.S. degree with the School of Physics and Electronics, Hunan University, Changsha, China. His current research interests are the circuit design of memristors and its application.
\end{IEEEbiography}
\vspace{-10pt}

\begin{IEEEbiography}[{\includegraphics[width=1in,height=1.25in,clip,keepaspectratio]{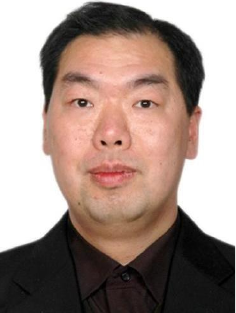}}]{Sichun Du}
 received the M.S degree and Ph.D degree in computer science and technology from Hunan University, Changsha, China, in 2005 and 2012, respectively. He is currently an Associate Professor with college of computer science and electronic engineering, Hunan University, Changsha 410082, China. His current research interests include memristive neural network, analog/RF integrated circuit synthesis and evolutionary computation algorithms.
\end{IEEEbiography}
\vspace{-510pt}

\begin{IEEEbiography}[{\includegraphics[width=1in,height=1.25in,clip,keepaspectratio]{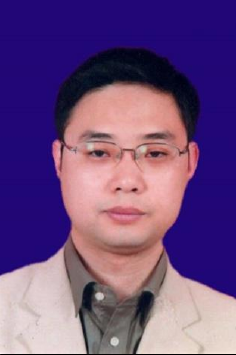}}]{Tao Li}
 received the M.Sc. degree in power electronics and transmission from Wuhan University, Wuhan, China, in 2001, and the Ph.D. degree in electrical engineering from Hunan University, Changsha, China, in 2011. Since 2003, he has been with the School of Computer Science and Electronic Engineering, Hunan University, where he is currently an Associate Professor.
His research interests include memristive neural network, artificial intelligence, neural network, and smart grid.
\end{IEEEbiography}

\end{document}